\documentclass[letterpaper, 10 pt, conference]{ieeeconf}  % Comment this line out if you need a4paper

\IEEEoverridecommandlockouts                              % This command is only needed if 
\usepackage{graphicx}
\usepackage{amsmath}
\usepackage{amsfonts}
\usepackage{nicefrac}
\usepackage{xcolor}
\usepackage{comment}
\usepackage{hyperref}
\usepackage{xspace}
\usepackage{multirow}
\usepackage{subcaption}

\usepackage{verbatimbox}

\let\labelindent\relax
\usepackage{enumitem}

\usepackage{blindtext}
\usepackage{graphbox}
\usepackage{textcomp}

\usepackage{amssymb}% http://ctan.org/pkg/amssymb
\usepackage{pifont}% http://ctan.org/pkg/pifont
\newcommand{\cmark}{\ding{51}}%
\newcommand{\xmark}{\ding{55}}%

\newcommand{\field}[1]{\mathbb{#1}}

\newcommand{\R}{\field{R}}

\newcommand{\bX}{\mathbf{X}}

\newcommand{\bs}{\mathbf{s}}
\newcommand{\bS}{\mathbf{S}}
\newcommand{\bt}{\mathbf{t}}

\newcommand{\by}{\boldsymbol{y}}
\newcommand{\bY}{\mathbf{Y}}

\newcommand{\mL}{\mathcal{L}}

\newcommand{\mB}{\mathcal{B}}
\newcommand{\mE}{\mathcal{E}}

\makeatletter
\DeclareRobustCommand\onedot{\futurelet\@let@token\@onedot}
\def\@onedot{\ifx\@let@token.\else.\null\fi\xspace}
\def\eg{\emph{e.g}\onedot} 
\def\ie{\emph{i.e}\onedot}

\makeatother

\newcommand{\gr}[1]{{\color[HTML]{BDBDBD}#1}}
\usepackage{hhline}

\title{\LARGE \bf
PaintNet: Unstructured Multi-Path Learning\\from 3D Point Clouds for Robotic Spray Painting
}

\author{Gabriele Tiboni$^{1}$ and Raffaello Camoriano$^{1}$ and Tatiana Tommasi$^{1}$% <-this % stops a space
\thanks{*This work was supported by the EFORT group, providing the authors with domain knowledge, original object meshes, trajectory data, and access to the proprietary spray painting simulator used during the experiments. 
We also acknowledge the support of the European H2020 Elise project (\url{www.elise-ai.eu}, and the CINECA award under the ISCRA initiative (DRE-URL - HP10CF881L) for the availability of HPC resources and support.
This study was carried out within the FAIR - Future Artificial Intelligence Research and received funding from the European Union Next-GenerationEU (PIANO NAZIONALE DI RIPRESA E RESILIENZA (PNRR) – MISSIONE 4 COMPONENTE 2, INVESTIMENTO 1.3 – D.D. 1555 11/10/2022, PE00000013). This manuscript reflects only the authors’ views and opinions, neither the European Union nor the European Commission can be considered responsible for them.
}% <-this % stops a space
\thanks{$^{1}$Politecnico di Torino, Turin, Italy {\tt\small first.last@polito.it}}%
}

\begin{document}

\maketitle
\thispagestyle{empty}
\pagestyle{empty}

%%%%%%%%%%%%%%%%%%%%%%%%%%%%%%%%%%%%%%%%%%%%%%%%%%%%%%%%%%%%%%%%%%%%%%%%%%%%%%%%
\begin{abstract}
Popular industrial robotic problems such as spray painting and welding require (i) conditioning on free-shape 3D objects and (ii) planning of multiple trajectories to solve the task. Yet, existing solutions make strong assumptions on the form of input surfaces and the nature of output paths, resulting in limited approaches unable to cope with real-data variability.
By leveraging on recent advances in 3D deep learning, we introduce a novel framework capable of dealing with arbitrary 3D surfaces, and handling a variable number of unordered output paths (\ie unstructured).
Our approach predicts local path segments, which can be later concatenated to reconstruct long-horizon paths. We extensively validate the proposed method in the context of robotic spray painting by releasing PaintNet, the first public dataset of expert demonstrations on free-shape 3D objects collected in a real industrial scenario.
A thorough experimental analysis demonstrates the capabilities of our model to promptly predict smooth output paths that cover up to 95\%
of previously unseen object surfaces, even without explicitly optimizing for paint coverage.
 
\end{abstract}
%%%%%%%%%%%%%%%%%%%%%%%%%%%%%%%%%%%%%%%%%%%%%%%%%%%%%%%%%%%%%%%%%%%%%%%%%%%%%%%%

\section{INTRODUCTION}
\label{sec:introduction}

\begin{table*}[t]
\caption{Literature review.}
\resizebox{\linewidth}{!}{%
\begin{tabular}{|c|c|c|c|c|c|c||c|c|c|c|c|}
\hline
\multirow{4}{*}{Applications} & \multirow{4}{*}{Works} & \multicolumn{2}{c|}{Input}& \multicolumn{3}{c||}{Output} &  \multirow{4}{*}{Method} &   \multicolumn{3}{c|}{Pros (+) and Cons (-)} \\ \cline{3-7}\cline{9-11}
& & Convex Objects & Highly  & \multirow{1.5}{*}{Single} & \multirow{1.5}{*}{Multi} & Unstructured &  & {Fast Path} & {Ability to} & \multirow{3}{*}{Others} \\
& & or Low-curvature  & Concave  & \multirow{1.5}{*}{Path} & \multirow{1.5}{*}{Path} & (unknown length,  &  & {Generation} & {Generalize} & \\
& & Surfaces &  Objects &  & &  num. of paths) & & (+) & (+) & \\
\hline
& \cite{Sheng_Automated_2000}\cite{Biegelbauer_Inverse_2005}\cite{Chen_Automated_2008}\cite{Li_Automatic_2010} & \cmark & \gr{\xmark}
&  \cmark & \cmark  & \cmark
& \multirow{3}{*}{Heuristics} & \multirow{3}{*}{\gr{\xmark}} & \multirow{3}{*}{\gr{\xmark}} & \multirow{3}{*}{\shortstack{(+) High painting coverage\\(-) High design cost and manual tuning}} \\ \cline{2-7}
{} & \cite{atkar24uniform}\cite{Andulkar_Incremental_2015}\cite{gleeson2022generating}   & \cmark  & \gr{\xmark} & \cmark & \gr{\xmark} & \gr{\xmark} & & & &     \\ \cline{2-7}
{Spray Painting}& \cite{Chen_Trajectory_2020}   & \cmark & \gr{\xmark} & \cmark & \cmark & \gr{\xmark}  & & & &\\ \cline{2-11}
%Painting & \cite{Biegelbauer_Inverse_2005} & \cmark  & \gr{\xmark} &  \cmark& \cmark& \cmark&  &  & &  \\ \cline{2-11}
%& \cite{Gasparetto_Automatic_2012} & \cmark & \gr{\xmark} &  \cmark& \cmark& \cmark&  &  & &  \\ \cline{2-11}
%\cite{Gasparetto_Automatic_2012}
& \multirow{2}{*}{\cite{Kiemel_PaintRL_2019}} & \multirow{2}{*}{\cmark} & \multirow{2}{*}{\gr{\xmark}} & \multirow{2}{*}{\cmark} &  \multirow{2}{*}{\gr{\xmark}}  &  \multirow{2}{*}{\gr{\xmark}}  & Reinforcement& \multirow{2}{*}\cmark& \multirow{2}{*}{\gr{\xmark}} &  (+) Explicit painting coverage optimization\\ 
&   & &   &   &    &     &  Learning  &   &  & (-) Requires an accurate simulator \\   \hline\hline 
%%%%%%%%%%%%%%%%%%%%%%
\multirow{4}{*}{\shortstack{Autonomous\\3D Inspection}} & 
\multirow{3}{*}{\cite{Englot_Hover_2012}\cite{Bircher2016ThreedimensionalCP}} & 
\multirow{3}{*}{\cmark} & 
\multirow{3}{*}{\cmark} &  
\multirow{3}{*}{\cmark} & 
\multirow{3}{*}{\gr{\xmark}} & 
\multirow{3}{*}{\gr{\xmark}} & 
\multirow{4}{*}{\shortstack{Coverage\\ Path Planning}} & 
\multirow{4}{*}{\gr{\xmark}} & 
\multirow{4}{*}{\gr{\xmark}} & 

\multirow{4}{*}{\shortstack{(+) High inspection coverage\\ (-) Sample-specific hyperparameters\\(-) Unable to model painting patterns}}
\\[3ex] \cline{2-7} %[10ex]%\hhline{--}
%%%%%%%%%%%%%%%%%%%%%%
 %& \cite{jing2020multi}\cite{multiUAV_2023}& \cmark & \cmark &  \cmark& \cmark & \gr{\xmark} &  &  & & \\ 
& 
\multirow{2}{*}{\cite{jing2020multi}\cite{multiUAV_2023}}  &  
\multirow{2}{*}{\cmark}&   
\multirow{2}{*}{\cmark}&   
\multirow{2}{*}{\cmark}&  
\multirow{2}{*}{\cmark}&   
\multirow{2}{*}{\gr{\xmark}}&  
&  
& 
& \\ 
 & &  &   &   &  &   &  &  & & \\ 
 %%%%%%%%%%%%%%%%%%%%
\hline
Programming by & \cite{NIPS2016_gail}\cite{behav_cloning}  & \gr{\xmark} & \gr{\xmark} & \cmark & \gr{\xmark} & \gr{\xmark} &  & \multirow{5}{*}{\cmark} & \multirow{5}{*}{\cmark} &  \\
\cline{2-7}
Demonstration & \cite{SRINIVASAN2021598} & \gr{\xmark} & \gr{\xmark} & \cmark & \cmark & \gr{\xmark} & \multirow{3}{*}{\shortstack{Imitation and\\ Supervised Learning}} &  & & \multirow{3}{*}{\shortstack{(+) Learns painting  patterns from data\\(-) Implicit paint coverage optimization}}\\
\cline{1-7} 
3D Deep Learning & \cite{Yuan_Pcn_2018}\cite{Alliegro_Denoise_2021} & \cmark &  \cmark &  \multicolumn{3}{|c||}{$\pmb{\sim}$ \ (point-wise predictions)}
&  &  & & {}\\
\cline{1-1}\hhline{~======~|~|~|~|}
%\multirow{2}{*}{Learning} & \cite{dexpoint} & & &  & & &  RL & High & \xmark & \\ \cline{5-11}
\multicolumn{1}{c}{\multirow{2}{*}{}} & \multicolumn{1}{|c|}{\multirow{2}{*}{Ours}}& \multirow{2}{*}{\cmark} & \multirow{2}{*}{\cmark} &  \multirow{2}{*}{\cmark} &  \multirow{2}{*}{\cmark} &  \multirow{2}{*}{\cmark} &   &  & & \\
\multicolumn{1}{c}{}    &  \multicolumn{1}{|c|}{}  & &   &    &    &    &    &   & & \\
\cline{2-11}
\end{tabular}%
}
\label{tab:literature_review}
\vspace{-12pt}
\end{table*}

Conditioning tasks on free-shape 3D objects is central to many industrial robotic applications, from grasping and manipulation to spray painting, welding and cleaning. Among them, all the tasks that unfold over long-time horizons require considerable amounts of computational resources for optimization and planning. Their key challenges are dealing with the inherent complexity of free-shape 3D input and with a high dimensional output that describes the full robot program. This scenario has led practitioners to introduce task-specific prior knowledge and data-specific simplifying assumptions.
Robotic spray painting is a representative example of this problem setting, where the robot must generate multiple trajectories for painting a surface, with each trajectory being a separate path through space. Even a simple planar surface becomes tricky if we consider both its two sides, and the difficulties grow when facing an object composed by convex and concave parts with different samples showing significant variability in shape and size. Clearly the number and length of output paths will differ for every instance, and will further change among object categories. Given the lack of affordable and flexible solutions, robotic spray painting remains a largely unsolved problem despite its relevance for product manufacturing.

\begin{figure}[tb]
    \centering
    \includegraphics[width=0.95\linewidth]{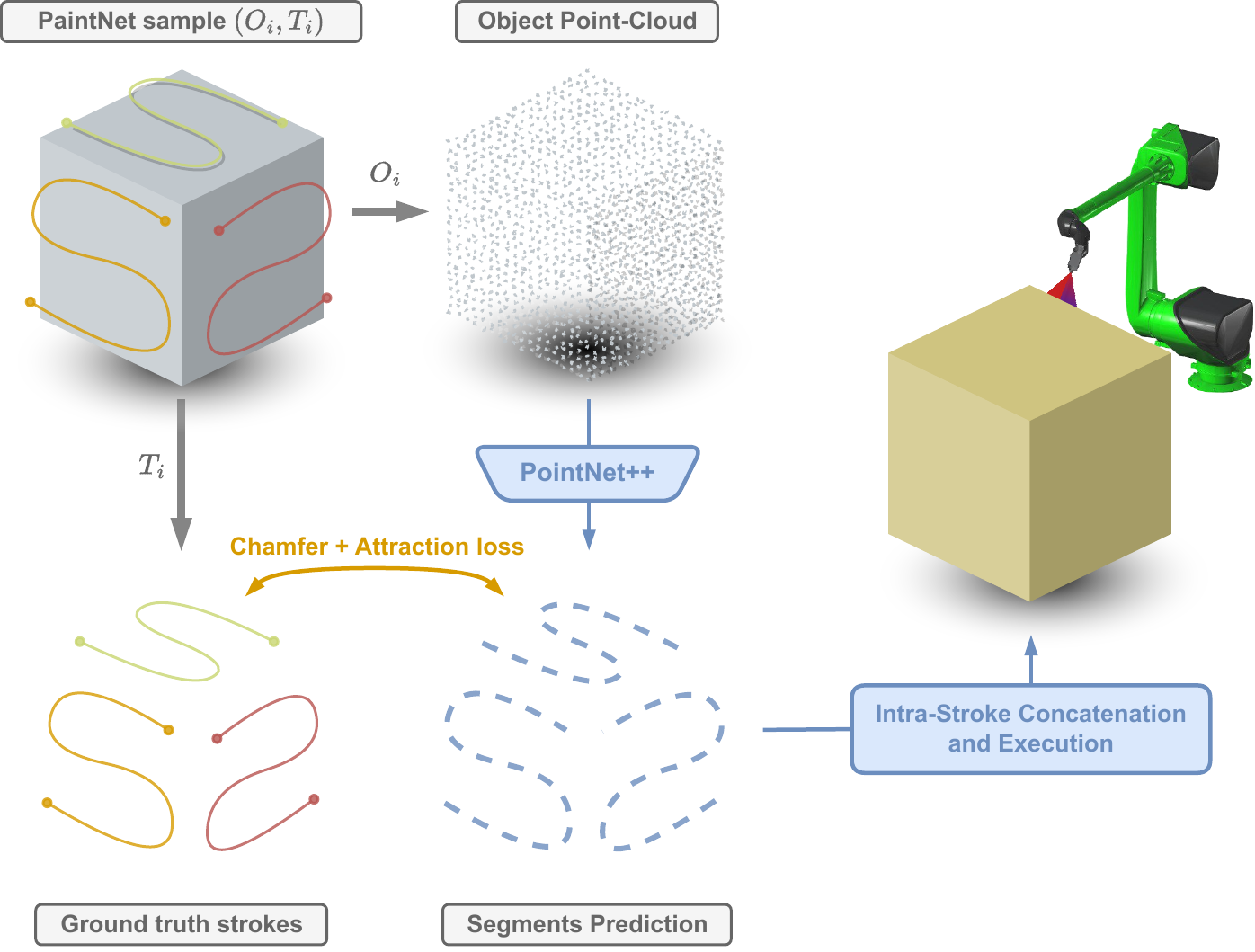}
    \caption{Overview of our method for multi-path prediction of 6D-pose spray painting paths given a raw 3D point-cloud.} \vspace{-5.5mm}
    \label{fig:intro_overview}
\end{figure}

Existing research studies rely on decoupling the task in (i) 3D object partitioning into convex surfaces and (ii) offline trajectory optimization through either domain-specific heuristics~\cite{Sheng_Automated_2000,Chen_Automated_2008,Li_Automatic_2010,Andulkar_Incremental_2015,atkar24uniform,gleeson2022generating}, or reinforcement learning-based policies~\cite{Kiemel_PaintRL_2019}.
Such approaches rely on simplified premises and are heavily tailored for specific shapes and convex surfaces only, which significantly restricts their ability to generalize to novel objects. 
Additionally, they often require expensive offline optimization routines, which hinder their practical applicability for industrial production lines. These limitations highlight the need for more suitable solutions that can operate on arbitrary 3D surfaces and efficiently handle complex multi-path planning problems.

In this work, we propose a novel method to address these challenges by designing a deep learning framework able to deal with unstructured high-dimensional input, such as 3D objects in the form of point clouds, and inherently cope with multiple and unordered output paths. Our approach learns a latent representation of a 3D object, and consecutively predicts local path segments that can be concatenated to reconstruct long-horizon robotic paths (see Fig.~\ref{fig:intro_overview}). Unlike heuristic methods that need to be re-designed ad hoc for every task and object, our framework can be applied to any 3D object-conditioned multi-path robotic task. 
Our data-driven approach only requires a set of expert demonstrations to learn from, and will remain effective and efficient regardless of the complexity of the object surfaces and the number of output paths.
We denote the output nature as \textit{unstructured}, as outputs paths are assumed to be unordered and variable in length and number.
An extensive validation of the proposed method is then presented in the context of robotic spray painting.
Overall, we present four main contributions:
\begin{itemize}[leftmargin=*]
    \item We introduce PaintNet, the first 3D object dataset annotated with expert spray painting demonstrations in a multi-path setting. PaintNet was collected in a real-world industrial scenario and includes a total of 845 samples, each defined by an object shape and its associated complex trajectory patterns.
    \item We design a novel learning-based framework 
    able to operate on free-shape 3D input and unstructured output paths. Our method predicts local path segments which are then concatenated to reconstruct long-horizon paths.
    \item We define a reproducible experimental benchmark with quantitative and qualitative metrics. We compare our approach with a baseline that directly regresses high-dimensional paths, and with a model that outputs separate point-wise 6D poses rather than segments. We show that the proposed method can effectively predict paths on previously unseen object instances in real-time, achieving up to 95\% spray painting coverage.
    \item Finally, we provide evidence on how the learned models can be leveraged when facing new object categories, improving performance at low data cost as well as speeding up convergence.
\end{itemize}

\section{RELATED WORK}
\label{sec:related_work}
In the following, we review existing literature in path planning for robotic spray painting and related fields, as well as learning-based approaches for path generation.
For a schematic overview, also refer to Table~\ref{tab:literature_review}.

\noindent\textbf{Planning for robotic spray painting.} 
Automatic robotic spray painting is an instance of the NP-hard coverage path planning (CPP) problem with additional complexity arising from the nonlinear dynamics of paint deposition and hard-to-model engineering experience acquired via trial and error by trajectory designers.
Due to its complexity, the landscape of robotic painting is dominated by \emph{heuristic} methods operating under simplifying assumptions about object geometry and generated path structure---\eg, raster patterns only. 
Critically, all existing heuristics assume to work with convex or low-curvature surfaces~\cite{Sheng_Automated_2000,Chen_Automated_2008,Li_Automatic_2010,Andulkar_Incremental_2015,atkar24uniform,gleeson2022generating}.
This renders them inapplicable to painting concave objects such as containers, for which more complex path patterns are required.
More recently, \cite{gleeson2022generating}~proposed a method to optimize painting quality by adapting trajectory waypoints and velocities. 
Still, it builds on an externally provided trajectory candidate and does not focus on long-horizon planning.
Besides the aforementioned  techniques, which require a 3D mesh or CAD model of the object,~\cite{Chen_Trajectory_2020} introduced a point cloud slicing procedure to compute global painting paths. 
However, this method is composed by multiple stages, each of which needs significant human expert guidance, and is still limited to simple convex objects. 
Other works rely on {matching} the objects with a combination of hand-designed elementary geometric components collected in a database~\cite{Biegelbauer_Inverse_2005}.
Matching components are associated with local painting strokes, which are then merged to form painting paths. Despite its merits, this method requires costly work by experts to explicitly codify object parts and their corresponding painting procedures for each object family.

\emph{Reinforcement learning} (RL) has also been proposed for training trajectory generators by directly optimizing a painting coverage reward~\cite{Kiemel_PaintRL_2019}, although limited to planar domains.
RL for stroke sequencing also proved successful for reconstructing 2D images~\cite{Huang_Learning_2019}. 
Although promising, RL is yet to be demonstrated successful for long-horizon 3D object planning due to the high dimensionality of the state and action spaces.
The need of an accurate simulator and low generalization of RL agents to novel objects also stand out as major issues.

We remark that all the mentioned works show results on a few proprietary object instances. 
They do not release either the data or the method implementation to allow a fair benchmark in terms of coverage performance and computational complexity, besides lacking a discussion on generalization to new object instances and categories which makes them practically ineffective. 

\noindent{\textbf{Autonomous 3D inspection.}}
As outlined above, planning for autonomous robotic painting is largely unsolved.
Interestingly, related CPP problems are being investigated in the setting of 3D inspection path planning for autonomous vehicles, which has several key features in common, \ie,
(i)~long-horizon mission paths, 
(ii)~concave objects to be inspected, and 
(iii)~joint planning of multiple paths may be required for teams of autonomous vehicles.
Iterative sampling and optimization methods 
have been presented for AUVs in~\cite{Englot_Hover_2012} and UAVs in~\cite{Bircher2016ThreedimensionalCP}.
More recently,~\cite{jing2020multi,multiUAV_2023} proposed optimization methods for planning multiple paths and demonstrated their effectiveness for multi-UAV missions on large structures.
Despite being significantly more capable of generating long-horizon paths around complex 3D objects than current planning methods for spray painting, they target visual inspection so the coverage goal differs from that of painting. 
Moreover, their high computational cost and sample-specific hyperparameters render them inapplicable to small-batch settings in which swift path generation is pivotal.

\noindent{\textbf{Robot programming by demonstration. }}
The programming by demonstration (PbD) paradigm~\cite{billard2008robot} obviates explicit programming of robot trajectories that might be costly or unfeasible.
Among the earliest PbD  approaches, kinesthetic teaching or teleoperation and replay of the recorded trajectory allow experienced operators to guide the robot along the desired path to complete the task.
However, such methods are highly specific and do not generalize to new tasks or objects. 
More advanced PbD methods based on Imitation Learning (IL) 
achieve better generalization and entail most computational costs during offline training while enabling fast path prediction~\cite{behav_cloning,NIPS2016_gail}. 
Still, most IL methods only support single-path generation, with the recent exception of~\cite{SRINIVASAN2021598}, which is however not suitable for unstructured outputs (\ie unknown number of unordered paths) and has been demonstrated in 2D domains only.
Furthermore, applying current IL methods to object-centric tasks such as robotic painting is not straightforward, since conditioning their input to 3D data is an open problem.

\noindent{\textbf{3D Deep Learning. }}
Recently introduced 3D Deep Learning architectures 
apply predictive models to free-shape 3D data. 
In particular, an object can be described by a 3D point cloud: an unstructured set of points generally collected by dedicated sensors (\eg, laser scanners). 
Such architectures take point clouds as input to efficiently perform tasks such as object classification~\cite{Qi_Pointnet_2017} and segmentation~\cite{Qi_Pointnet++_2017}, shape completion~\cite{Yuan_Pcn_2018, Alliegro_Denoise_2021}, and robotic grasping, where the output is a structured grasp descriptor~\cite{Ni_Pointnet++Grasp_2020,Alliegro_End_2022}. 
In particular, shape completion architectures output 3D points that cover missing object regions, resulting in point-wise predictions that we remark could potentially be employed as path waypoints.
In this work, we leverage the expressive power and high-dimensional output capabilities of 3D Deep Learning architectures for shape completion. We adapt them to efficiently generate multiple long-horizon robotic spray painting strokes while generalizing to new object instances, thus overcoming the limitations of heuristics, costly optimization-based CPP methods, and classical PbD.

\section{THE PAINTNET DATASET}
\label{sec:dataset}
\begin{table}[tb]
\caption{Summary of the PaintNet dataset characteristics. Top: (a) a simple 3D object with $I=3$ painting strokes $\{\mathbf{t}\}_{i=1}^I$, each composed of a sequence of $N_i$ 6D  poses. (b) and (c) are closeups showing the 6D poses respectively for a window and a container. Bottom: dataset information per object category indicating their increasing complexity.}
\resizebox{\columnwidth}{!}{%
\begin{tabular}{|c|c|c|c|c|c|}
\multicolumn{2}{@{}c@{}}{\includegraphics[width=4cm]{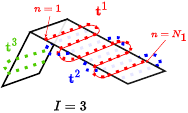}} &
\multicolumn{2}{c@{}}{\includegraphics[width=3cm]{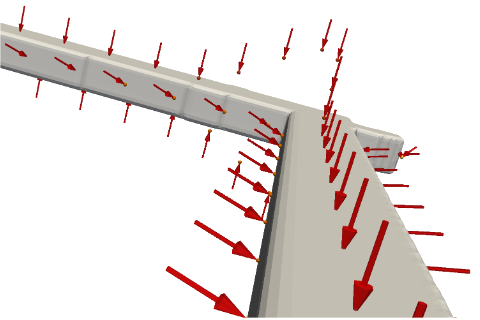}}& 
\multicolumn{2}{c}{\includegraphics[width=3cm]{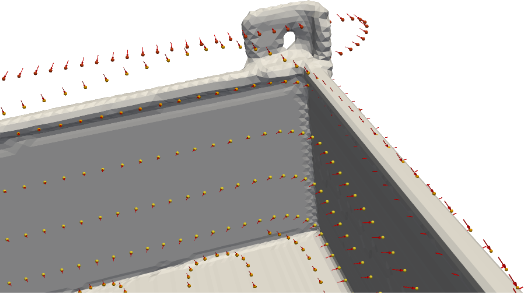}}\\
\multicolumn{2}{c}{(a)} & \multicolumn{2}{c}{(b)} & \multicolumn{2}{c}{(c)}\\
\hline
\multirow{1.5}{*}{\textbf{Object}} &  \multirow{1.5}{*}{\textbf{Number}} &
  \textbf{Number} &
  \multicolumn{3}{c|}{\textbf{Complexity}} \\ \cline{4-6} 
\multirow{1.5}{*}{\textbf{Categories}} & \multirow{1.5}{*}{\textbf{of samples}} & \textbf{of strokes} &  \textbf{Varying num.} & \textbf{Concave} & \textbf{High shape}\\
& & \textbf{per sample} & \textbf{of strokes} & \textbf{surfaces} & \textbf{diversity} \\
\hline
  
\textbf{Cuboids}    & 300 & 6     & \multicolumn{1}{c|}{}       & \multicolumn{1}{c|}{}       &        \\
\textbf{Windows}    & 145 & 10 $\pm$ 5  & \multicolumn{1}{c|}{\cmark} & \multicolumn{1}{c|}{\cmark}       &        \\
\textbf{Shelves}    & 312 & 20 $\pm$ 14 & \multicolumn{1}{c|}{\cmark} & \multicolumn{1}{c|}{\cmark} &        \\
\textbf{Containers} & 88  & 16 $\pm$ 5 & \multicolumn{1}{c|}{\cmark} & \multicolumn{1}{c|}{\cmark} & \cmark \\ \hline
\end{tabular}%
}
\label{tab:dataset_complexity}
\vspace{-12pt}
\end{table}
\begin{figure*}[tb]
    \centering
    \includegraphics[width=0.9\linewidth]{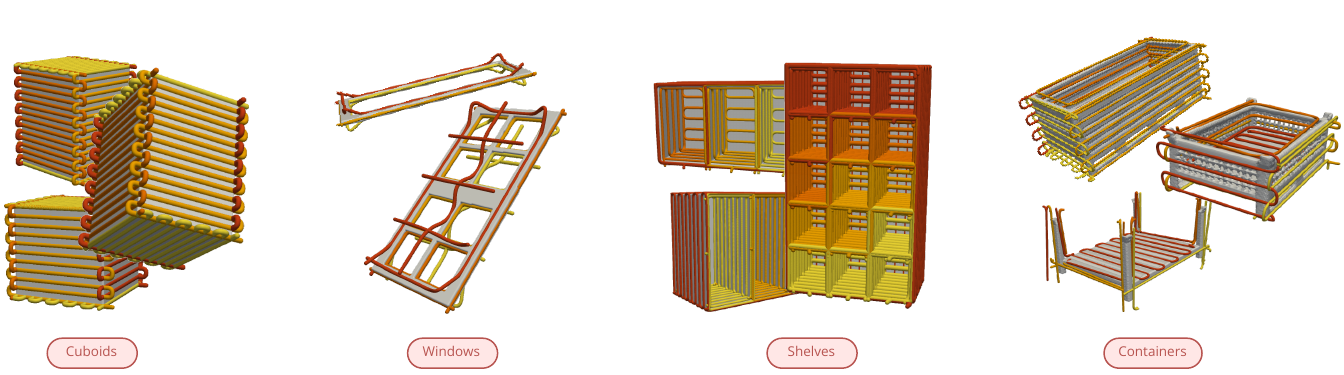}
    \caption{Overview of a few representative instances for each of the four categories included in the PaintNet dataset.}
    \label{fig:paintnet_samples_overview}
    \vspace{-12pt}
\end{figure*}

We introduce the PaintNet dataset with the aim of providing the community with a public testbed for multi-path prediction conditioned on free-shape 3D objects.
It is composed by $(O,T)$ samples which are pairs of an object shape $O$ and its corresponding spray painting paths set $T$.
Each object shape is a triangular mesh $O=(V,F)$ defined by vertices $V \in \R^{|V|\times 3}$ and faces $F$. The three coordinates of each vertex are expressed in real-world millimeter scale.
The paths set is formalized as a set of sequences $T=\{\bt^i\}_{i=1}^I$. Each sequence is referred to as a \textit{stroke}, varying in length and number across objects: $\bt^i$ encodes the spray painting gun position and orientation along the stroke, containing a variable number of poses $t^i_{n=1,\ldots,N_i} \in \R^6$. 
More precisely, we record positions (3D) as the ideal paint deposit point---12cm away from the gun nozzle---and gun orientations (3D) as Euler angles. 
Each pose is collected by sampling from the end-effector kinematics at a rate of 4ms during offline program execution.
An overview of the characteristics of the dataset is reported in Tab \ref{tab:dataset_complexity}, with a number of representative samples illustrated in Fig.~\ref{fig:paintnet_samples_overview}.
The four object categories composing the dataset are presented in the following, ordered by growing complexity.

\noindent\textbf{Cuboids}: a confined class of 300 rectangular cuboid-shaped objects which allows to test models under minimal generalization requirements and simpler path patterns.
Cuboids vary in height and depth, and are associated with six simple raster-like paths designed to paint the exterior faces.

\noindent\textbf{Windows}: a set of 145 window-like 3D meshes from real-world use cases, provided with their hand-crafted spray painting paths. In contrast with the previous class, windows introduce harder challenges for path generation, such as predicting a non-stationary number of strokes, and handling non-trivial gun orientations (\eg see Tab.~\ref{tab:dataset_complexity} (b)).

\noindent\textbf{Shelves}: a set of 312 shelf-like objects characterized by 
highly concave surfaces. 
A strategy dealing with separate surface patches would not be enough in this case, leading to unfeasible global patterns where the gun interferes with surrounding patches. 
Shelf meshes differ by volume size and number of inner shelves. 
Their associated ground truth paths have been generated by skilled practitioners through manually-defined rules.

\noindent\textbf{Containers}: a set of 88 industrial containers including meshes with various surface concavities and instances with fairly heterogeneous global and local (wavy and grated surfaces) geometric properties.
The related painting paths have been designed by experts and obtained through a manually-guided process which shows irregularities among samples.

The data was generously provided by the EFORT group\footnote{\url{https://efort.com.cn/en/index.php/group}} and later %manually 
preprocessed by the authors. In particular, all object meshes are released in a subdivided, aligned, and smoothed watertight~\cite{Huang_Robust_2018} version to avoid sharp edges and holes. Moreover, any private information (\eg, original logos) was accurately anonymized. 
The PaintNet dataset is publicly available at \url{https://gabrieletiboni.github.io/paintnet/}.

\section{METHOD}
\label{sec:method}
\subsection{Method Overview}
We approach multi-path learning for spray painting as a point cloud-based inference task, and present a tailored deep learning model to deal with unstructured output paths.

The input point cloud can be obtained by laser scanning the workpiece to be sprayed, which avoids the need for the exact CAD model from the object designer. When the object mesh is available, as in our case, the point cloud is simply generated by sampling from the known surface, \eg through Poisson Disk sampling~\cite{Cook_Stochastic_1986}.
To address learning of an unknown number of output paths, we design the model output as a set of path segments, which are intended to be subsets of the original strokes. The fixed length $\lambda \in \mathbb{N}^+$ of each segment is a hyperparameter of the model. An optimal trade-off between the number and length of predicted path segments can inherently cope with the varying number of unordered strokes and varying stroke lengths.
By the same logic, ground-truth paths are also decomposed in $\lambda$-length segments and used as a reference for the training process. The final objective of our deep learning model consists in predicting path segments that are smoothly aligned with one another and can be concatenated to resemble the original strokes.

\subsection{Segments Prediction}
We denote the set of path segments as $\bS=\{\bs^k\}_{k=1}^{K}$.
Each segment is composed of $\lambda$ ordered poses obtained from the ground truth strokes, with $\bs^k \in \R^{\lambda \times 6}$. Specifically, we consider an overlap of one pose among consecutive within-stroke segments to encourage contiguous predictions, resulting in a total number of $K=\sum_{i=1,\ldots,I}\lfloor \nicefrac{N_i-\lambda}{\lambda-1}\rfloor+1$ ground-truth segments.

Our model takes as input the object point cloud $\bX$ composed of unordered 3D points $x_{p=1,\ldots,P} \in \R^3$, and provides as output a set of path segments $\bY=\{\by^k\}_{k=1}^{K^*}$ with $\by^k \in \R^{\lambda \times 6}$, and $y^k_l \in \R^6$ denoting the $l$-th pose element in the $k$-th segment.
Considering that the value of $K$ may slightly vary with the instances, we set $K^*=\lfloor \nicefrac{(\sum_{i=1,\ldots,I}N_i)-\lambda}{\lambda-1}\rfloor+1$ as upper bound to fit them all. Other pre-processing techniques might be adopted to ensure a fixed number of ground-truth path segments.
The learning objective is pursued by minimizing the following loss: 
\begin{equation}
    \mL_{y2s} = \frac{1}{K^*}\sum_{\by\in \bY} \min_{\bs \in \bS}\|\by-\bs\|_2^2 + \frac{1}{K}\sum_{\bs \in \bS} \min_{\by\in \bY}\|\bs-\by\|_2^2~.
    \label{eq1}
\end{equation}
In words, this symmetric version of the Chamfer Distance~\cite{Fan_Point_2017} drives the prediction of permutation-invariant path segments close to the ones in the ground truth.

We leverage the partial overlap of within-stroke ground-truth segments to furtherly encourage contiguous path segments in space and drive a similar behavior in the model predictions. This will also facilitate the concatenation of generated segments at the post-processing stage (see Sec.~\ref{sec:intra_stroke_alignment}).
To this end, we introduce two sets of poses $\mB=\{y^k_1\}_{k=1}^{K^*}$ and $\mE=\{y^k_\lambda\}_{k=1}^{K^*}$, that respectively collect the beginning and ending poses of predicted segments.
We then introduce an additional Chamfer-based loss which guides segments to have overlapping initial and ending poses:
\begin{equation}
    \mL_{b2e} = \frac{1}{2K^*}\Big\{\sum_{y^k_1\in \mB} \min_{y^j_\lambda \in \mE}\|y^k_1-y^j_\lambda\|_2^2 + \sum_{y^k_\lambda \in \mE} \min_{y^j_1 \in \mB}\|y^k_\lambda -y^j_1\|_2^2 \Big\}~,
\end{equation}
with  $j\neq k$. 
Overall, we train our model to optimize $\mL = \mL_{y2s} + \alpha\mL_{b2e}$, with $\alpha \in \mathbb{R^+}$.

\subsection{Intra-stroke Concatenation}
\label{sec:intra_stroke_alignment}
Although the execution of unordered path segments would be theoretically feasible in unstructured multi-path settings, in practice this may lead to problematic cycle times on real hardware.
To this end, we note that post-processing steps may be adopted to order intra-stroke segments, \eg, through domain knowledge or ad-hoc heuristics by paint specialists. More advanced solutions may include a combination of segment clustering and the solution of the TSP problem on each cluster.
Here we show that a simple technique based on segment proximity and alignment may be just as effective to link predicted segments into long-horizon paths.

Specifically, we interpret the segments as nodes of a graph and we 
aim at concatenating them such that each segment $k$ has at most one outgoing $e^{+}_k\leq 1$, and one incoming edge $e^{-}_k \leq 1$, where $e$ is the signed edge degree.
For each segment $k$, we evaluate the distance
$d_{k} =min_{j} \| y_{\lambda }^k -y_{1}^{j} \| _2^2 +\| ( y_{\lambda }^{k} -y_{\lambda -1}^k) -( y_2^j -y_1^j) \| _2^2$
s.t. $j\neq k$ and $e^{-}_j=0$,
which considers proximity in space and orientation, as well as similarity in segment directions. 
Then, we connect two segments with a directed edge from $k$ to $j$ in case $d_{k}$ falls below a defined threshold $\tau$, proceeding in ascending order of $d_k$.

Finally, we merge via averaging the ending and beginning poses of the two segments at hand, leveraging the redundant overlapping poses induced in the training process.
The single hyperparameter $\tau$ can be selected to achieve desired stroke reconstruction while preserving spray painting coverage (see Section~\ref{sec:results_intra_stroke}).

\section{EVALUATION METRICS}
\label{sec:evaluation_metrics}
To fairly assess the performance of our approach and the considered baselines we introduce two evaluation metrics.

\noindent\textbf{Pose-wise Chamfer Distance (PCD)~\cite{Fan_Point_2017}}. 
It compares the predicted and ground truth paths as two clouds of 6D-poses.
This metric accounts for the predicted gun positions and orientations, while disregarding the structured nature of the predictions, \ie the intra-segment connectivity.

\noindent\textbf{Paint Coverage (PC)}. Although not directly optimized at training time, we wish to assess the percentage of surface covered by the predicted strokes when executed on a spray painting simulator, relative to the ground truth. 
We start by defining a per-mesh painting thickness threshold above which a vertex is identified as covered: we set it as the $10_{th}$ percentile of non-zero ground-truth thickness values for the mesh in question.
Then, on the subset of covered ground-truth vertices, we evaluate the percentage of vertices covered when executing our predicted strokes.
Note how this metric is independent of the specific spray gun model parameters used during simulation (\eg paint flux), thus it is suitable for benchmarking purposes.

\section{EXPERIMENTS}
\label{sec:experiments}
\begin{table*}[tb]
\caption{\textbf{Left}: Predicted poses on representative PaintNet test instances (light blue) and the ground-truth strokes (orange). \textbf{Right}: Spray painting coverage visualization when executing predicted and expert poses on a spray painting simulator. The colormap ranges from green (low) to yellow (high).}\label{tab:qualitative}
\resizebox{\textwidth}{!}{
\begin{tabular}{c|c@{}c@{}c@{}c|c|}
\cline{2-6}
& Point-Wise & Multi-Path & Ours & \textbf{Ours} & \multirow{2}{*}{Ground Truth}\\
& Prediction & Regression & ($\lambda=10$) & ($\lambda=4$) & \\
 \hline
\multicolumn{1}{|@{}c@{}|}{\rotatebox[origin=c]{90}{Cuboids}} &
\includegraphics[align=c,height=1.8cm]{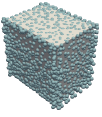} &
\includegraphics[align=c,height=1.65cm]{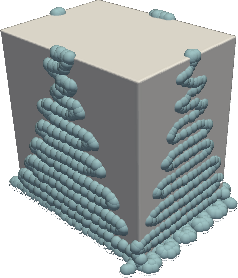} &
\includegraphics[align=c,height=1.8cm]{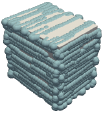} & 
\includegraphics[align=c,height=1.8cm]{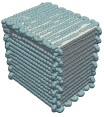} & 
\includegraphics[align=c,height=1.8cm]{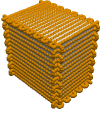}\\
 \hline
 \multicolumn{1}{|@{}c@{}|}{\rotatebox[origin=c]{90}{Windows}} &
 \includegraphics[align=c,height=1.8cm]{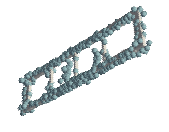} &  \includegraphics[align=c,height=0.6cm]{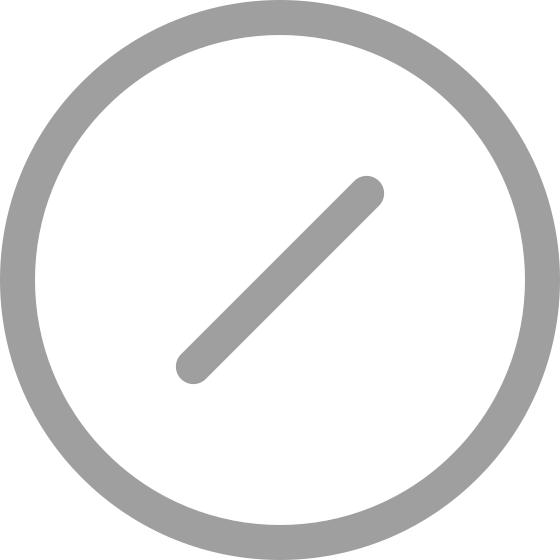} &
 \includegraphics[align=c,height=1.8cm]{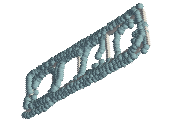} &
  \includegraphics[align=c,height=1.8cm]{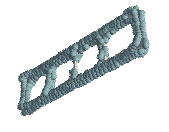} & 
 \includegraphics[align=c,height=1.8cm]{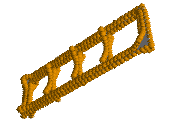}\\
 \hline
 \multicolumn{1}{|@{}c@{}|}{\rotatebox[origin=c]{90}{Shelves}} &
 \includegraphics[align=c,height=1.8cm]{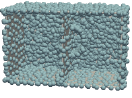} & \includegraphics[align=c,height=0.6cm]{02_images/slash_v2.png} &
 \includegraphics[align=c,height=1.8cm]{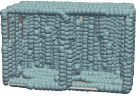} &
  \includegraphics[align=c,height=1.8cm]{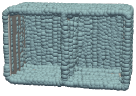} & 
 \includegraphics[align=c,height=1.8cm]{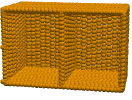}\\
 \hline
 \multicolumn{1}{|@{}c@{}|}{\rotatebox[origin=c]{90}{Containers}} &  
 \includegraphics[align=c,height=1.8cm]{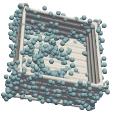}  & 
 \includegraphics[align=c,height=0.6cm]{02_images/slash_v2.png} &
 \includegraphics[align=c,height=1.8cm]{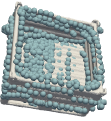} &
  \includegraphics[align=c,height=1.8cm]{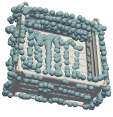} & 
 \includegraphics[align=c,height=1.8cm]{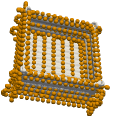}\\
 \hline
\end{tabular}
 \begin{tabular}{|c@{}c@{}c@{}c|c|}
 \hline
Point-Wise & Multi-Path & Ours & \textbf{Ours} & \multirow{2}{*}{Ground Truth}\\
Prediction & Regression & ($\lambda=10$) & ($\lambda=4$) & \\
\hline
\includegraphics[align=c,height=1.8cm]{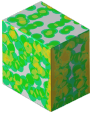} &
\includegraphics[align=c,height=1.7cm]{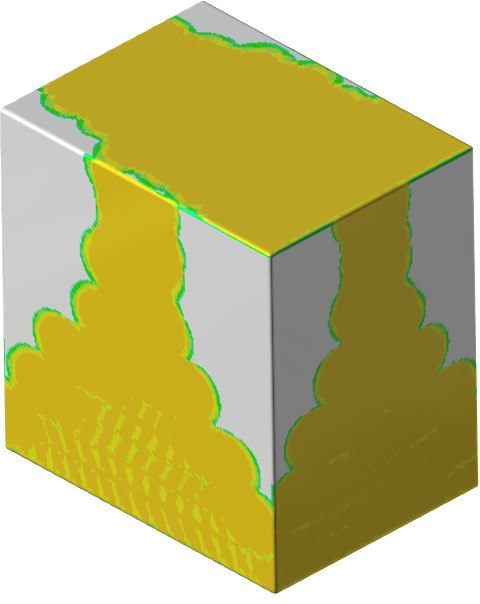}& 
\includegraphics[align=c,height=1.8cm]{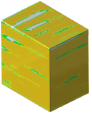}& 
\includegraphics[align=c,height=1.8cm]{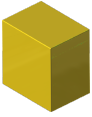}& 
\includegraphics[align=c,height=1.8cm]{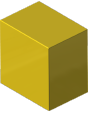}\\
\hline
\includegraphics[align=c,height=1.8cm]{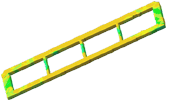} & \includegraphics[align=c,height=0.6cm]{02_images/slash_v2.png} &
\includegraphics[align=c,height=1.8cm]{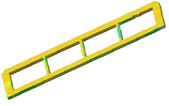}& 
\includegraphics[align=c,height=1.8cm]{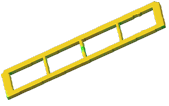}&
\includegraphics[align=c,height=1.8cm]{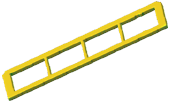}\\
\hline
\includegraphics[align=c,height=1.8cm]{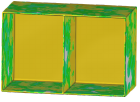} & \includegraphics[align=c,height=0.6cm]{02_images/slash_v2.png} &
\includegraphics[align=c,height=1.8cm]{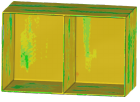} & 
\includegraphics[align=c,height=1.8cm]{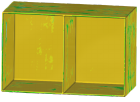} & 
\includegraphics[align=c,height=1.8cm]{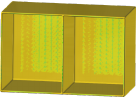}\\
\hline
\includegraphics[align=c,height=1.8cm]{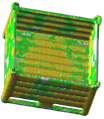} & \includegraphics[align=c,height=0.6cm]{02_images/slash_v2.png} &
\includegraphics[align=c,height=1.8cm]{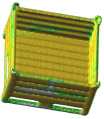} & 
\includegraphics[align=c,height=1.8cm]{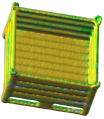} & 
\includegraphics[align=c,height=1.8cm]{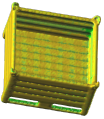}\\
 \hline
 \end{tabular}
}
\vspace{-12pt}
\end{table*}

\subsection{Implementation Details}
Our pipeline leverages an encoder architecture based on PointNet++~\cite{Qi_Pointnet++_2017}, that acts as a feature extractor from the input point cloud of 5120 3D points to a latent space of dimensionality 1024. A 3-layer MLP is then appended to generate output poses, with hidden size (1024,1024) and output size $(\lambda \times 6) \times K^*$.
We encode the orientation of output 6D poses as a 3D unit vector---rather than Euler angles---by applying an L2-normalization to
%the dedicated 3 output neurons.
the 3 output neurons corresponding to the orientation components of the predicted pose.
Consequently, ground truth Euler Angles are converted into 2-DoF 3D unit vectors and used as ground-truth in~(\ref{eq1}), effectively penalising predicted orientations according to cosine similarity.
This is permitted by our conic spray gun model, which is invariant to rotations around the approach axis. 
A weight vector is introduced when computing the distance between $\by$ and $\bs$ 
to properly combine location and orientation.
Overall, we optimize our loss function $\mathcal{L}$ with $\alpha=0.5$,
orientation vectors weighted by $0.25$,
learning rate $10^{-3}$, Adam optimizer, and $1200$ epochs.

To cope with limited training data, we initialize our network with pre-trained weights from a shape classification task on ModelNet~\cite{wu_modelnet_2014}.
Input point clouds and ground truth paths are normalized during training by independently centering to zero mean and down-scaling by a category-specific factor.
Rather than directly dealing with poses densely sampled every 4ms, we down-sample expert trajectories to a number of poses
$L=\{2000, 500, 4000, 1000\}$ respectively for \{\emph{cuboids}, \emph{windows}, \emph{shelves}, \emph{containers}\} (or $L=2000$ for join-training experiments in Sec.~\ref{subsec:generalization}).
Finally, we randomly split each category into training-test sets with 80\%-20\% respective proportions. 
All results reported in the manuscript are computed on previously unseen test instances.

\noindent\textbf{Baselines.} As discussed in Sec.~\ref{sec:related_work}, the complex task at hand that combines the challenges of free-shape 3D objects as input and unstructured output paths has not been faced by previous literature.
Therefore, we design two baselines tailored to our setting.
One is a deep learning model inspired by shape completion methods~\cite{Yuan_Pcn_2018, Alliegro_Denoise_2021}, that outputs 6D poses instead of 3D points. We indicate it as \emph{point-wise prediction} since it shares the same architecture as our method but ignores connectivity of output poses, resembling the particular case of $\lambda=1$.
In the attempt to preserve some structure in the output space, we develop a second variant that regresses complete output strokes rather than single poses, referred to as \textit{multi-path regression}. However, this comes at the cost of fixing the number and length of output strokes a priori, resulting in a reference approach suitable only for the \emph{cuboids} category.

\begin{figure}[tb]
    \centering
    \begin{subfigure}{0.35\linewidth}
    \centering
    \includegraphics[width=\linewidth]{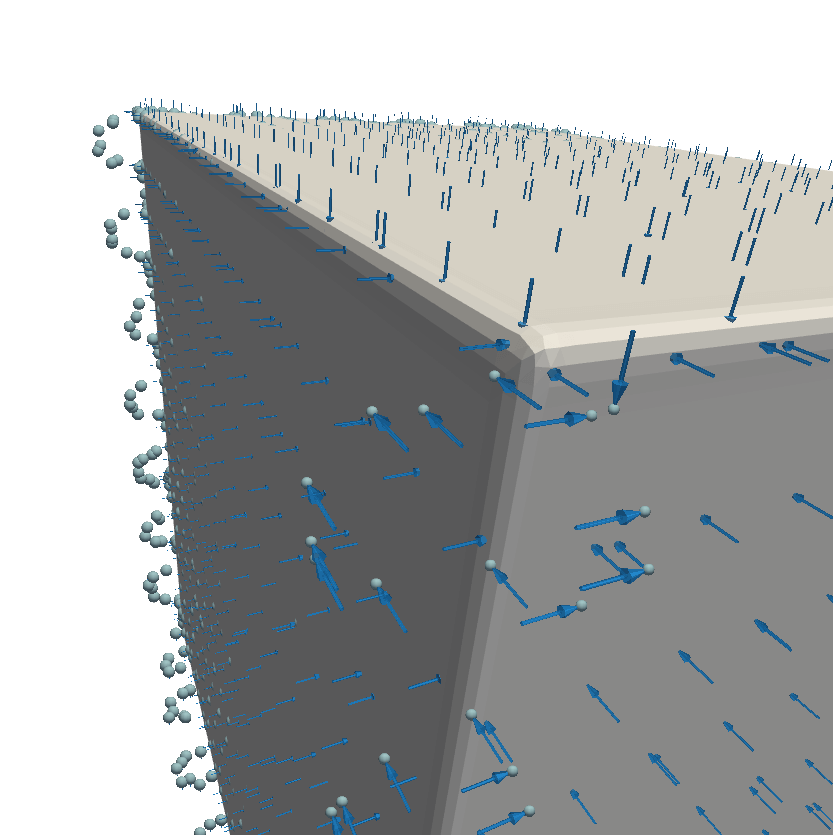}
    \caption{}
    \label{fig:sub1}
    \end{subfigure}
    \begin{subfigure}{0.35\linewidth}
    \centering
    \includegraphics[width=\linewidth]{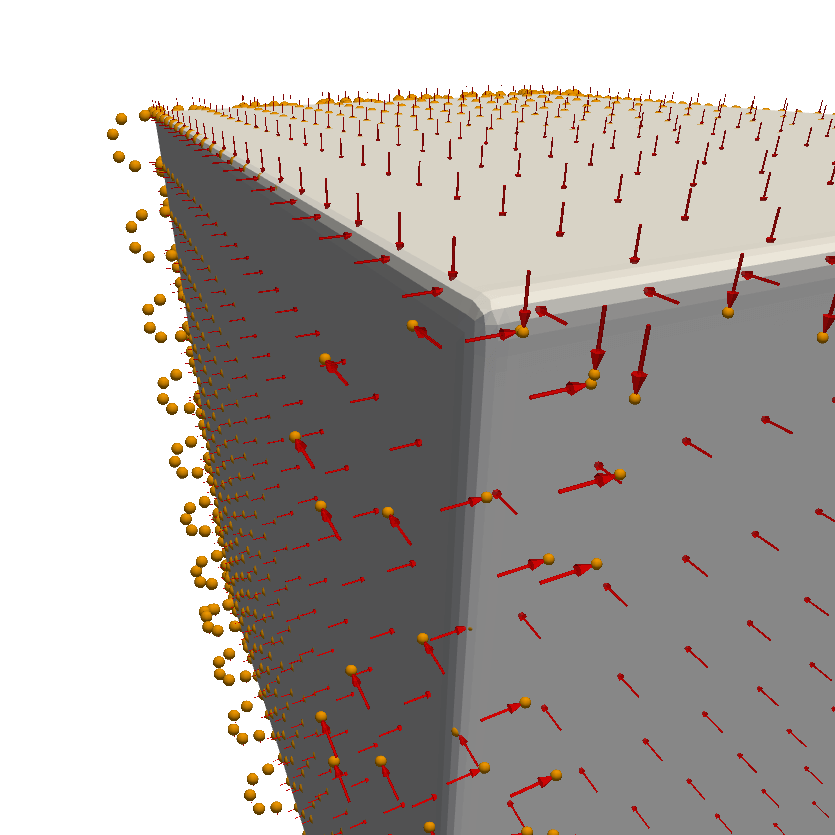}
    \caption{}
    \label{fig:sub2}
    \end{subfigure}
    \vspace{-6pt}
    \caption{(\subref{fig:sub1}) Predicted vs. (\subref{fig:sub2}) ground-truth 6D-poses ($\lambda=4$). Pose orientations are efficiently preserved and learned.}
    \label{fig:normals_learned}
    \vspace{-16pt}
\end{figure}

\begin{table}
\centering
\caption{Chamfer Distance averaged over each category's test set, up-scaled by $10^4$. The lower the better.
%The number of total output poses $L$ is reported in parenthesis.
}
\def\arraystretch{1.25}%  1 is the default, change whatever you need
\resizebox{\columnwidth}{!}{%
\begin{tabular}{l|c|c|c|c|}
\cline{2-5}
                                   & 
\def\arraystretch{1}\begin{tabular}[c]{@{}c@{}}Cuboids\\ %($L=2000$)
\end{tabular} & 
\def\arraystretch{1}\begin{tabular}[c]{@{}c@{}}Windows\\ %($L=500$)
\end{tabular} & 
\def\arraystretch{1}\begin{tabular}[c]{@{}c@{}}Shelves\\ %($L=4000$)
\end{tabular} & 
\def\arraystretch{1}\begin{tabular}[c]{@{}c@{}}Containers\\ %($L=1000$)
\end{tabular} \\ \hline
\multicolumn{1}{|l|}{Point-Wise Prediction}  & 959.29                                                   & 950.72                                                  & 455.74                                                   & 1073.15                                                    \\ \hline
\multicolumn{1}{|l|}{Multi-Path Regression} & $8.32\times10^5$                                                    & -                                                  & -                                                    & -                                                      \\ \hline
\multicolumn{1}{|l|}{Ours ($\lambda=10$)} & 37.98                                                    & 118.50                                                  & 56.06                                                    & 364.54                                                      \\
\multicolumn{1}{|l|}{Ours ($\lambda=4$)}  & \textbf{18.25}                                           & \textbf{57.17}                                          & \textbf{36.65}                                           & \textbf{274.84}                                             \\ \hline
\end{tabular}
}
%\vspace{-3pt}
\label{tab:quantitative_chamfer}
% \gab{Here, $\delta$ refers to the overlapping parameter, we'll talk about this tomorrow but it's likely that we'll skip introducing the attraction loss altogether, in favor of simply introducing the overlapping as it already does the job pretty nicely.}}
\vspace{-6mm}
\end{table}

\subsection{Results: Segments Prediction}
\label{subsec:segments_prediction}
As the PaintNet dataset comes with four different categories of varying complexity and structure, in this section we carry out separate trainings for each category, while keeping the same hyperparameters.
This already provides hints on the robustness of our pipeline on multiple object categories.
We report qualitative results on a subset of test instances in Tab.~\ref{tab:qualitative} (Left), and the full quantitative results on the test set in terms of PCD in Tab.~\ref{tab:quantitative_chamfer}.
Despite optimizing for the PCD evaluation metric explicitly, the %6D-pose shape completion 
point-wise prediction 
baseline applied to path generation leads to highly sparse poses, failing to preserve structure across all object categories. We attribute this shortcoming to the inherent nature of the Chamfer Distance used: despite able to deal %while it can deal 
with permutation-invariant data, it does not encourage predictions to be contiguous.
Interestingly enough, directly regressing a known number of $6$ strokes of length $333$ for the cuboids category also turns out to be problematic due to compounding errors on euclidean distances among high-dimensional pose sequences. % of 6D-poses.
As an intermediate case we also consider our approach with $\lambda=10$ which shares similar issues with the multi-path regression, failing to capture long-spanning patterns. 
We conclude that the long-horizon nature of the output strokes is just as critical to take into account when approaching the task, and may not simply be learned with na\"ive regression techniques.
On the other hand, with $\lambda=4$ we observe the capability of our model to predict output path segments that closely resemble the ground truth and maintain a contiguous structure across all categories.
Intuitively, the network is biased towards learning local spray painting patterns, which drastically simplifies the task and does not require learning implicit high-level planning. At the same time, the attraction loss term $\mathcal{L}_{b2e}$ assures aligned and  contiguous predictions with nearby segments.
We report a close-up illustration of predicted poses vs. ground-truth poses in Fig.~\ref{fig:normals_learned}, demonstrating successful learning of both positions and 2-DoF orientations, even when different strokes locally intersect each other.
Finally, we observe the limited capacity of our model to produce high-quality looking strokes on the most complex \emph{containers} category. This set of data introduces challenges such as high shape heterogeneity, complex spiral trajectory patterns, and fewer available training samples (70). We claim that leveraging additional data from similar tasks could offer a promising avenue for mitigating these difficulties, as evidenced in Sec.~\ref{subsec:generalization}. 

\begin{table}[tb]
\caption{Spray painting coverage: percentage of covered mesh vertices with respect to ground-truth trajectories. Results are averaged over the test set. The higher the better.}
\resizebox{\columnwidth}{!}{%
\centering
\def\arraystretch{1.25}%  1 is the default, change whatever you need
\begin{tabular}{l|c|c|c|c|}
\cline{2-5}
                                   & 
\def\arraystretch{1}\begin{tabular}[c]{@{}c@{}}Cuboids\\ %($L=2000$)
\end{tabular} & 
\def\arraystretch{1}\begin{tabular}[c]{@{}c@{}}Windows\\ %($L=500$)
\end{tabular} & 
\def\arraystretch{1}\begin{tabular}[c]{@{}c@{}}Shelves\\ %($L=4000$)
\end{tabular} & 
\def\arraystretch{1}\begin{tabular}[c]{@{}c@{}}Containers\\ %($L=1000$)
\end{tabular} \\ \hline
\multicolumn{1}{|l|}{Point-Wise Prediction}  & 5.42\%                                                      & 39.90\%                                                     & 26.40\%                                                      & 71.99\%                                                         \\ \hline
\multicolumn{1}{|l|}{Multi-Path Regression} & 79.41\%                                                      & -                                                     & -                                                     & -                                                         \\ \hline
\multicolumn{1}{|l|}{Ours ($\lambda=10$)} & 79.64\%                                                      & 68.84\%                                                     & 70.88\%                                                      & 82.88\%                                                         \\
\multicolumn{1}{|l|}{Ours ($\lambda=4$)}  & \textbf{95.30\%}                                                      & \textbf{84.05\%}                                                     & \textbf{73.03}\%                                                      & \textbf{89.32}\%                                                         \\ \hline
\end{tabular}
}
\vspace{-3pt}
\label{tab:quantitative_paintcoverage}
\vspace{-16pt}
\end{table}

\subsection{Results: Spray Painting Coverage}
As intra-stroke concatenation may lead to over-optimistic coverage percentages due to wrongly connected sequences, we first perform a thorough paint coverage analysis on the sole, disconnected segments. Note how, even though predictions lack inter-sequence connections at this stage, the overlapping component allows---at least in theory---a smooth spray gun transition from one sequence to another, without skipping steps.
We therefore obtain a painting feedback by executing each predicted segment in simulation in a random permutation. On the other hand, ground-truth paint thickness references are obtained through the execution of the known long-horizon trajectory. A proprietary simulator developed by the EFORT group is used for this step, but similar tools may equally serve the scope~\cite{Andulkar_Incremental_2015}. Qualitative results on a few instances of PaintNet are depicted in Table~\ref{tab:qualitative} (Right), with a color map design that matches our paint coverage metric defined in Sec.~\ref{sec:evaluation_metrics}, \ie vertex thicknesses higher than a relative threshold are considered to be covered and visually appear the same. Quantitative paint coverage values are reported in Table~\ref{tab:quantitative_paintcoverage}.
Overall, we draw similar conclusions as for the inference analysis: uniformly sparse poses predicted by the point-wise prediction model lead to poor coverage results, while the contiguous nature of predicted path segments with $\lambda=4$ allows for up to 95.30\% surface coverage and best overall coverage across all object categories. These results importantly demonstrate that supervised learning is a promising approach for learning the downstream task from expert data, even without directly optimizing for spray painting coverage.

\subsection{Results: Intra-stroke Concatenation}
\label{sec:results_intra_stroke}
\begin{figure}
\resizebox{0.5\textwidth}{!}{
\begin{tabular}{|c@{~}|@{~}c@{~}|@{~}c@{~}|@{~}c@{~}||@{~}c|}
\hline
\multicolumn{2}{|c|@{~}}{{{\xmark}} attraction, overlapping} & \multicolumn{2}{c||@{~}}{{{\cmark}} attraction, overlapping} & ground truth\\
\hline
{{\xmark}} \footnotesize{intra-stroke con.} & {{\cmark}} \footnotesize{intra-stroke con.} & {{\xmark}} \footnotesize{intra-stroke con.} & {{\cmark}} \footnotesize{intra-stroke con.} & \\
\includegraphics[align=c,height=1.8cm]{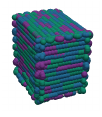} & \includegraphics[align=c,height=1.8cm]{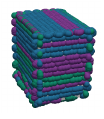}& 
\includegraphics[align=c,height=1.8cm]{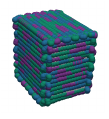} & \includegraphics[align=c,height=1.8cm]{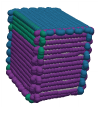} & 
\includegraphics[align=c,height=1.8cm]{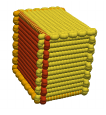}\\
 \hline
 PC = 75.06\% &  PC = 72.29\% & PC = 98.32\% & PC = 97.94\% & PC = 100\%\\
 \hline
 \includegraphics[align=c,height=1.8cm]{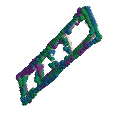} & \includegraphics[align=c,height=1.8cm]{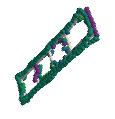}& 
\includegraphics[align=c,height=1.8cm]{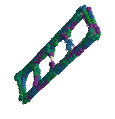} & \includegraphics[align=c,height=1.8cm]{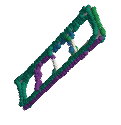} & 
\includegraphics[align=c,height=1.8cm]{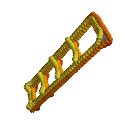}\\
\hline
PC = 87.70\% &  PC = 87.51\% & PC = 93.13\% & PC = 92.83\% & PC = 100\%\\
 \hline
\end{tabular}
}
\caption{Intra-stroke concatenation post-processing step ($\tau=0.15$) on cuboids and windows, from our approach ($\lambda=4$).}
\label{fig:intra_stroke_alignment}
\vspace{-12pt}
\end{figure}

We inspect the outcome of our proposed post-processing step in the attempt to reconstruct longer strokes for practical execution on robotic systems.
By design, our training pipeline already encourages the prediction of overlapping segments and allows a simple technique based on segment proximity to be applied, avoiding complex ordering procedures.
We demonstrate the effectiveness of the intra-stroke concatenation step in Fig.~\ref{fig:intra_stroke_alignment}, highlighting the contribution of both the attraction loss $\mL_{b2e}$ and overlapping component to obtain optimal qualitative and quantitative results.
We note that coverage results are preserved after the concatenation step, albeit not exactly the same: this effect is likely due to the merging of overlapping poses and smoothing.

\subsection{Results: Generalization}
\label{subsec:generalization}
Up to our knowledge, we are proposing the first unstructured multi-path prediction method formalized as a supervised learning task. This approach comes with some key advantages as the possibility to easily re-train the models when more data become available with no need for changing the architecture or re-engineering the process.
It is also possible to benefit from pre-trained models by obtaining reliable performance even in case of data and time constraints. 
These are realistic scenarios in industrial settings and in this section we investigate them, showing the generalization abilities of our approach by focusing on the most challenging \emph{containers} category.

\begin{figure}
\resizebox{0.5\textwidth}{!}{
\begin{tabular}{|cc|cc|}
\hline
\multicolumn{2}{|c|}{containers-specific model (L=2000)}   & \multicolumn{2}{c|}{joint-training (L=2000)} \\
\hline
\multirow{2}{*}[3em]{\includegraphics[width=0.25\linewidth]{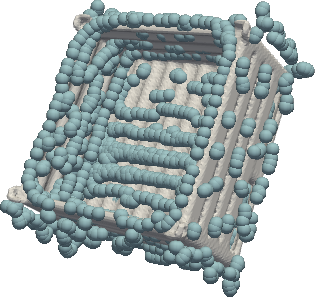}
} &
\includegraphics[width=0.25\linewidth]{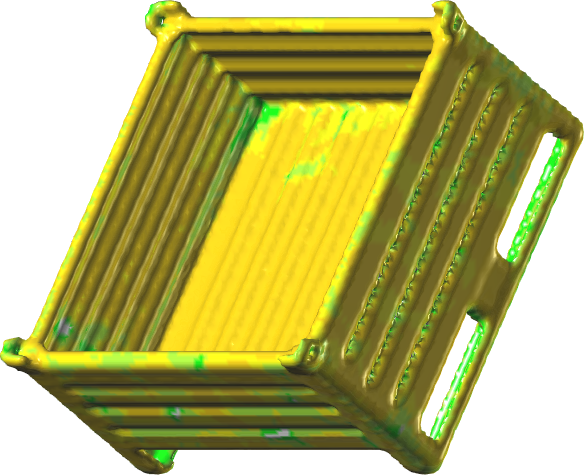} &\multirow{2}{*}[3em]{\includegraphics[width=0.25\linewidth]{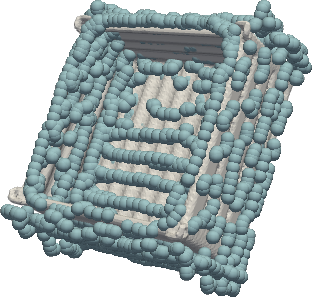}} & \includegraphics[width=0.25\linewidth]{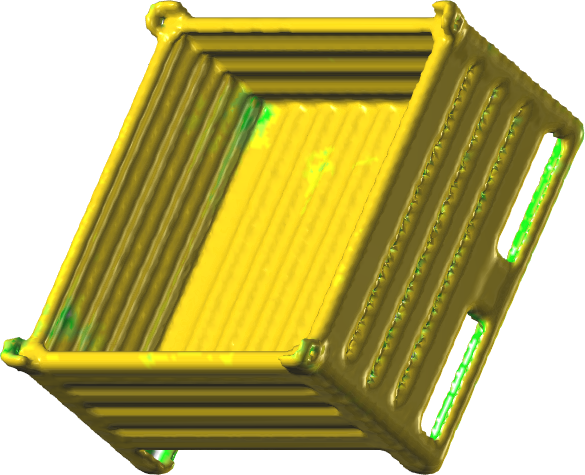} \\
\begin{tabular}[c]{@{}c@{}} PCD ($\times 10^4$): 232.60\\ PC: 89.93\%\end{tabular}
& \includegraphics[width=0.25\linewidth]{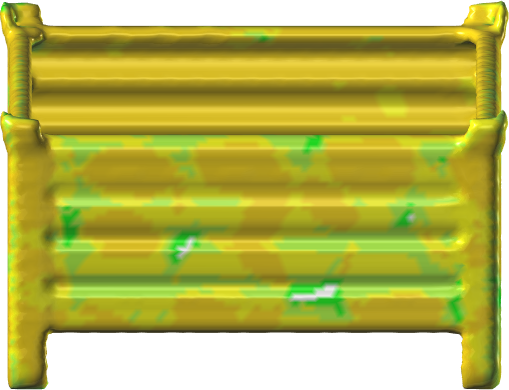} & 
\begin{tabular}[c]{@{}c@{}} PCD ($\times 10^4$): \textbf{164.61}\\ PC: \textbf{95.69\%}\end{tabular}
&\includegraphics[width=0.25\linewidth]{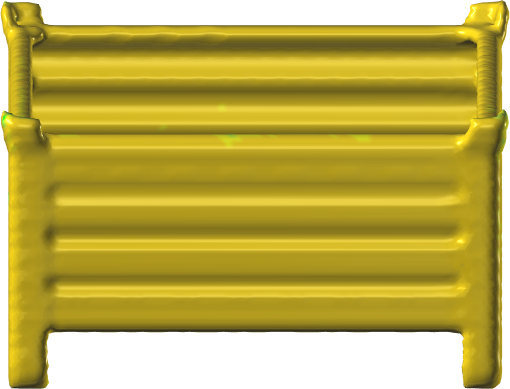}\\
\hline
\end{tabular}
}
    \caption{Qualitative example (predicted poses and respective coverage from two points of view) and performance comparison between the containers-specific model and the joint-training model when testing and averaging over all containers test instances. Colormap range: green (low), yellow (high). 
    }
    \label{fig:joint_training_containers}
\end{figure}

\noindent\textbf{Joint-training.}
We learned a model on all four object classes covered by the PaintNet dataset (whole PaintNet training set, 675 samples). This allows the network to observe a large variability in 3D shapes and painting patterns and better capture their relation. The comparison between the obtained performance and that of a model learned only on the containers training set (70 samples) is presented in Fig.~\ref{fig:joint_training_containers} and shows the benefit of leveraging additional data.  

\begin{figure}[tb]
    \centering
    \includegraphics[width=\linewidth]{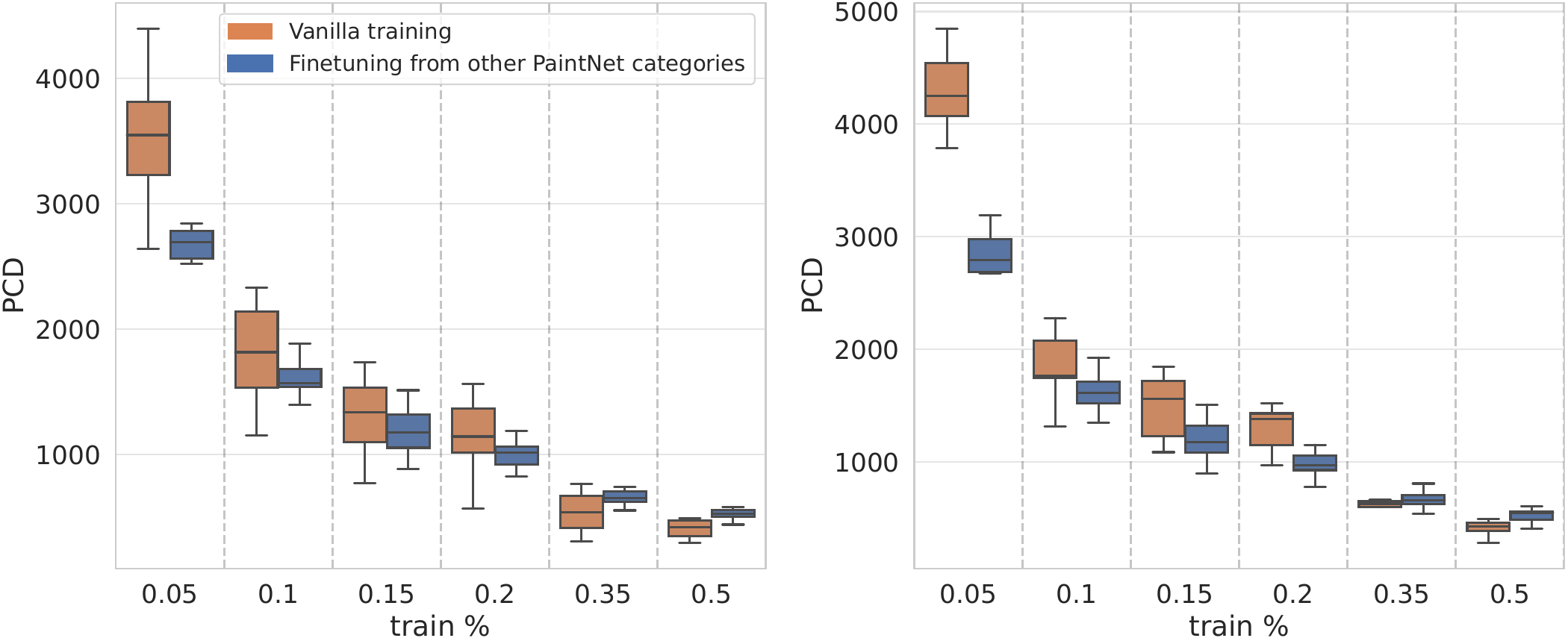}
    \caption{Few-shot: a model jointly pre-trained on cuboids, shelves, and windows
    generalizes better when finetuned on a subset of containers. Results on test set after (left) 600 and (right) 1200 training epochs; 12 repetitions.}
    \label{fig:few_shot_containers_boxplot}
    \vspace{-8pt}
\end{figure}

\noindent\textbf{Few-shot.} 
When only a very limited number of annotated samples is available for a new object class, a pre-trained model on related tasks may provide significant learning support. This behavior is perfectly exemplified by the results in Fig.~\ref{fig:few_shot_containers_boxplot} where we consider 5\% to 50\% subsets of the containers training set. 
We compare our vanilla training procedure on containers with a transfer one, that finetunes a model previously trained jointly on cuboids, windows, and shelves. 
The results confirm that the knowledge acquired from the other PaintNet categories can effectively be inherited to improve the performance on the containers, showing a marginal negative transfer in case of enough available training samples.

\begin{figure}[tb]
    \centering
    \includegraphics[width=\linewidth]{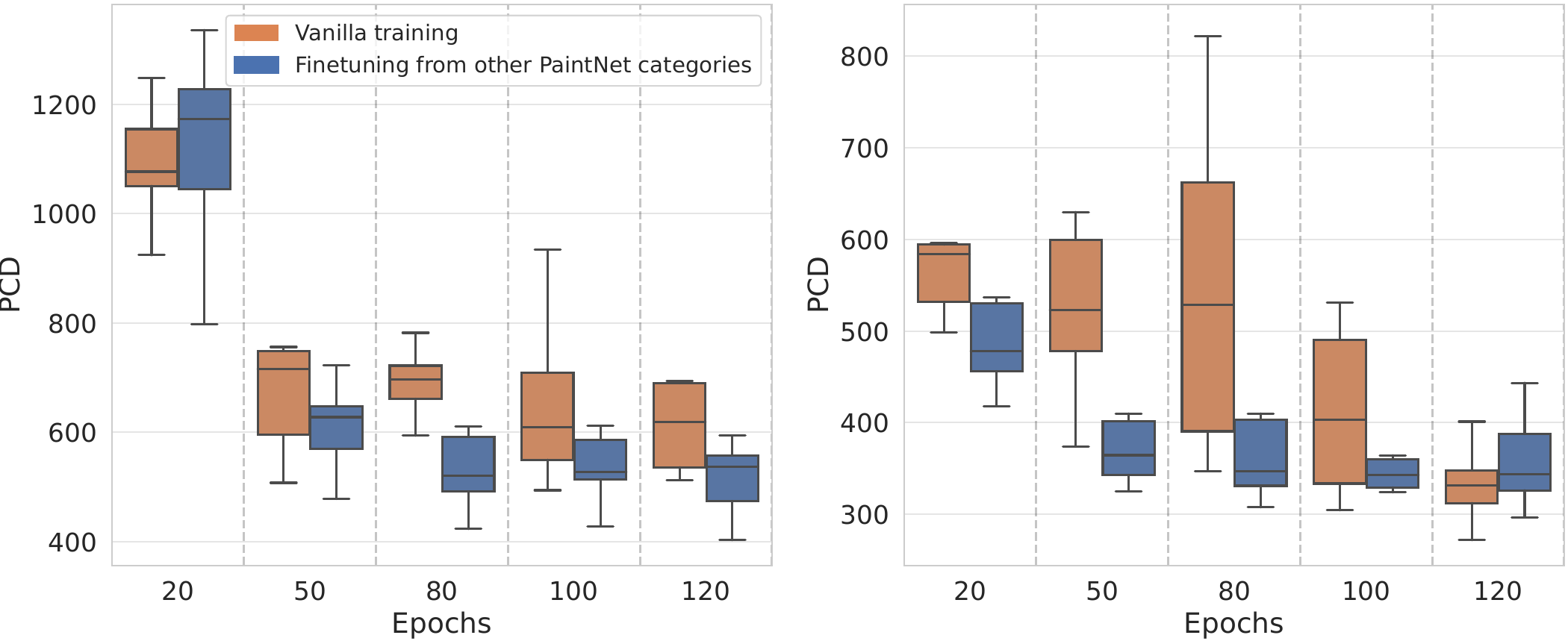}
    \vspace{-16pt}
    \caption{Convergence speed: a model jointly pre-trained on cuboids, shelves, and windows leads to faster convergence when finetuned on containers. Training with (left) 50\%  and (right) 100\% of available containers; 12 repetitions.}
    \label{fig:conv_speed_containers}
\end{figure}

\noindent\textbf{Convergence speed.} 
Analogous results to the few-shot case can be observed when the constraint is on the training time. %rather than on the amount of training data. 
Fig.~\ref{fig:conv_speed_containers} shows the effect of pre-training on the other PaintNet categories when the number of learning epochs on the containers is reduced by 90\% (from 1200 to 120). Finetuning the model learned from cuboids, windows and shelves leads to faster training convergence.  

In line with these findings, we encourage practitioners in the field to foresee the long-term potential of our supervised learning based model in future applications, as real-time inference capabilities combined with an increasing number of data may drastically reduce robot programming times. 

\subsection{Hyperparameters Sensitivity Analysis}
\begin{figure}[tb]
    \centering
    \vspace{-6pt}
    \includegraphics[width=0.8\linewidth]{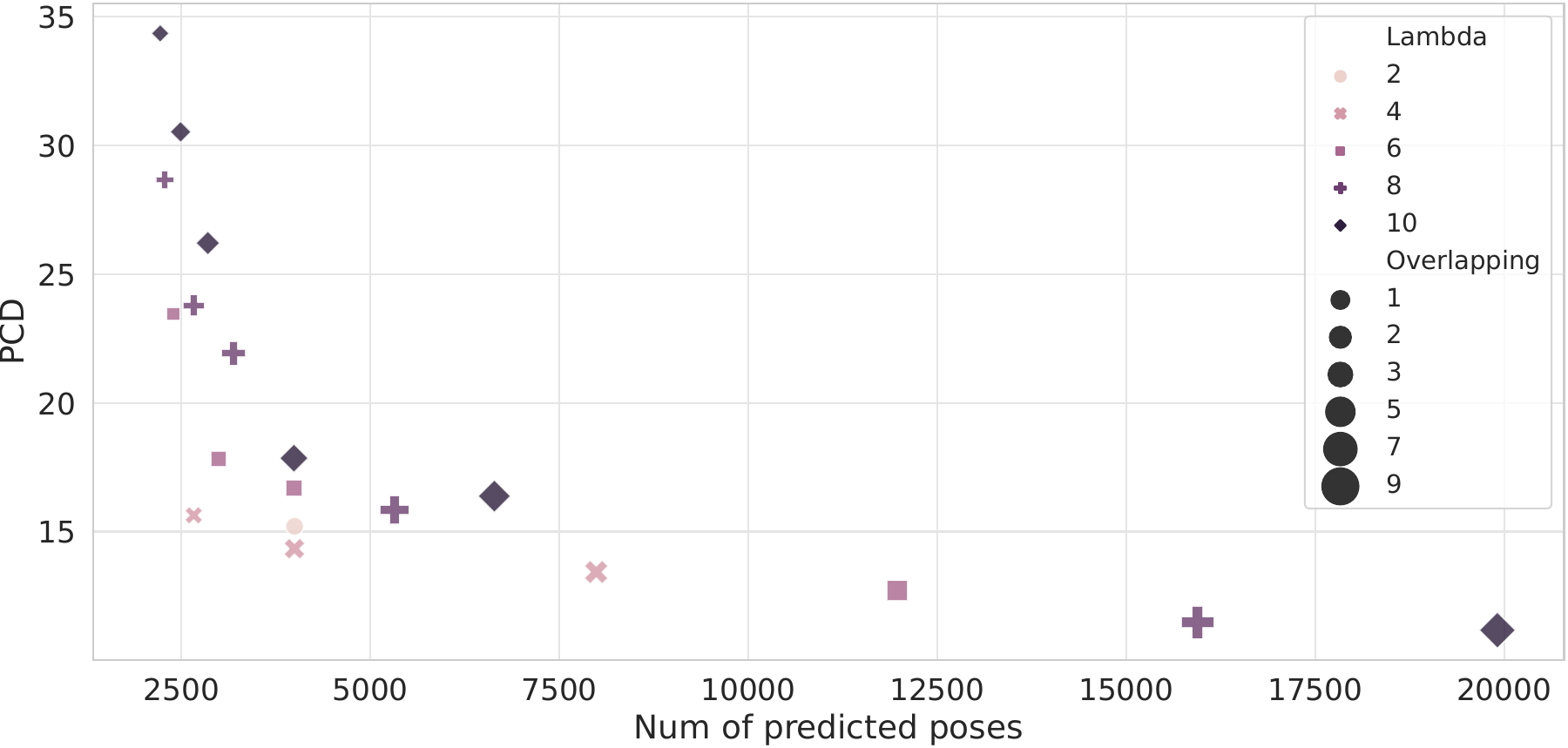}
    \vspace{-4pt}
    \caption{Sensitivity to lambda and overlapping parameters.}
    \label{fig:lambda_sensitivity}
    \vspace{-16pt}
\end{figure}

While $\lambda=4$ provided the best results across our experiments,  its value may be tuned according to the sampling frequency of output poses for the task at hand. We provide an illustration of the impact of the segment length $\lambda$ on the PCD metric in Fig.~\ref{fig:lambda_sensitivity}, for cuboids. The figure motivates our choice of lambda, as it reaches lower PCD values for the same number of predicted poses. 
We remark that increasing the intra-stroke overlap inevitably implies a growing number of poses: this causes the PCD to decrease due to points being naturally closer in space rather than to a real result improvement. Thus, a fair PCD comparison can be done only at a fixed number of predicted poses.

\section{CONCLUSIONS}
\vspace{-4pt}
\label{sec:conclusions}
In this paper, we tackle the core robotic problem of long-horizon, multiple path generation for tasks involving free-shape 3D objects.
To this aim, we focus on robotic spray painting as a particularly well-suited task in such domain.
In this context, we introduce PaintNet, the first industry-grade dataset for robotic spray painting, and present a novel supervised learning method to approach the underlying task via segments prediction and concatenation.
We validate the performance of our method in simulation, demonstrating promising paint coverage despite this metric not being optimized for explicitly.
Future work enabled by PaintNet will focus on real-world executability, \eg, by addressing semantically correct intra-stroke concatenation, mesh collision avoidance and predicted pose reachability.
Furthermore, incorporating painting quality optimization with complementary approaches~\cite{gleeson2022generating} may also lead to improved performance.
Finally, we believe the proposed approach can pave the way for further research on other long-horizon multi-path tasks in robotics conditioned on 3D objects (\eg, sanding, welding, or cleaning).

\addtolength{\textheight}{-4cm}   % This command serves to balance the column lengths

\bibliographystyle{IEEEtran}
\bibliography{root}
%%%%%%%%%%%%%%%%%%%%%%%%%%%%%%%%%%%%%%%%%%%%%%%%%%%%%%%%%%%%%%%%%%%%%%%%%%%%%%%%

\end{document}